\documentclass[sigconf, nonacm]{acmart} %,anonymous,review
\AtBeginDocument{%
  }

\usepackage[most]{tcolorbox} 
\usepackage{algorithmic}
\usepackage{graphicx}
\graphicspath{{./figs/}}
\DeclareGraphicsExtensions{.png,.pdf}
\usepackage{textcomp}
\usepackage{xcolor}
\usepackage{longtable}
\usepackage{booktabs}
\usepackage{tabularx}
\usepackage{array}

\DeclareRobustCommand{\hypobox}[1]{%
\begin{tcolorbox}[  
        breakable,
        left=0pt,
        right=0pt,
        top=0pt,
        bottom=0pt,
        width=\dimexpr\columnwidth\relax, 
        enlarge left by=0mm,
        boxsep=5pt,
        arc=3pt,outer arc=3pt,
        ]
        #1
\end{tcolorbox}
}

\newif\ifshowcommentshao
\newif\ifshowcommentsgopi
\newif\ifshowtodos

\showcommentsgopitrue     % Gopi comments + Jim replies
\showtodostrue            % TODO items

\newif\ifshowallcomments
\showallcommentsfalse
\ifshowallcomments
    \ifshowcommentshao
        \newcommand{\hao}[1]{\fcolorbox{gray}{yellow}{\bfseries\sffamily\scriptsize Hao}{\sf\small\textit{\textcolor{red}{#1}}}}
        \newcommand{\james}[1]{{\color{green!70!black}\textbf{[James:} #1\textbf{]}}}
    \else
        \newcommand{\hao}[1]{}
        \newcommand{\james}[1]{}
    \fi
\else
    \newcommand{\hao}[1]{}
    \newcommand{\james}[1]{}
\fi

\ifshowallcomments
    \ifshowcommentsgopi
        \newcommand{\gopi}[1]{\fcolorbox{gray}{cyan}{\bfseries\sffamily\scriptsize Gopi}{\sf\small\textit{\textcolor{blue}{#1}}}}
        \newcommand{\jim}[1]{{\color{green!70!black}\textbf{[Jim:} #1\textbf{]}}}
    \else
        \newcommand{\gopi}[1]{}
        \newcommand{\jim}[1]{}
    \fi
\else
    \newcommand{\gopi}[1]{}
    \newcommand{\jim}[1]{}
\fi

\ifshowallcomments
    \ifshowtodos
        \newcommand{\todo}[1]{{\color{red}\textbf{[TODO:} #1\textbf{]}}}
    \else
        \newcommand{\todo}[1]{}
    \fi
\else
    \newcommand{\todo}[1]{}
\fi

\renewcommand\footnotetextcopyrightpermission[1]{}
\acmConference{}{}{}

\begin{document}

%%
%% The "title" command has an optional parameter,
%% allowing the author to define a "short title" to be used in page headers.
\title{Permissive-Washing in the Open AI Supply Chain: A Large-Scale Audit of License Integrity}

%%
%% The "author" command and its associated commands are used to define
%% the authors and their affiliations.
%% Of note is the shared affiliation of the first two authors, and the
%% "authornote" and "authornotemark" commands
%% used to denote shared contribution to the research.
\author{James Jewitt}
\affiliation{%
  \institution{Queen's School of Computing, Queen's University}
  \city{Kingston}
  \state{Ontario}
  \country{Canada}}
\email{james.jewitt@queensu.ca}

\author{Gopi Krishnan Rajbahadur}
\affiliation{%
  \institution{Queen's School of Computing, Queen's University}
  \city{Kingston}
  \state{Ontario}
  \country{Canada}}
\email{grajbahadur@acm.org}

\author{Hao Li}
\affiliation{%
  \institution{Queen's School of Computing, Queen's University}
  \city{Kingston}
  \state{Ontario}
  \country{Canada}}
\email{hao.li@queensu.ca}

\author{Bram Adams}
\affiliation{%
  \institution{Queen's School of Computing, Queen's University}
  \city{Kingston}
  \state{Ontario}
  \country{Canada}}
\email{bram.adams@queensu.ca}

\author{Ahmed E. Hassan}
\affiliation{%
  \institution{Queen's School of Computing, Queen's University}
  \city{Kingston}
  \state{Ontario}
  \country{Canada}}
\email{ahmed@cs.queensu.ca}

%%
%% By default, the full list of authors will be used in the page
%% headers. Often, this list is too long, and will overlap
%% other information printed in the page headers. This command allows
%% the author to define a more concise list
%% of authors' names for this purpose.
\renewcommand{\shortauthors}{Jewitt et al.}

\newcommand{\rqone}{Do permissively-labeled assets (MIT, Apache-2.0, BSD-3-Clause) satisfy their mandatory legal requirements?}
\newcommand{\rqtwo}{For permissively-labeled assets (MIT, Apache-2.0, BSD-3-Clause), is upstream copyright attribution preserved across supply chain transitions?}

%%
%% The abstract is a short summary of the work to be presented in the
%% article.
\begin{abstract}

% \gopi{Needs to be updated to reflect all the changes I have made}

Permissive licenses like MIT, Apache-2.0, and BSD-3-Clause dominate open-source AI, signaling that artifacts like
models, datasets, and code can be freely used, modified, and redistributed. However, these licenses carry mandatory
requirements: include the full license text, provide a copyright notice, and preserve upstream attribution, that
remain unverified at scale. Failure to meet these conditions can place reuse outside the scope of the license,
effectively leaving AI artifacts under default copyright for those uses and exposing downstream users to litigation.
We call this phenomenon ``permissive washing'': labeling AI artifacts as free to use, while omitting the legal
documentation required to make that label actionable. To assess how widespread permissive washing is in the AI supply
chain, we empirically audit 124,278 dataset $\rightarrow$ model $\rightarrow$ application supply chains, spanning
3,338 datasets, 6,664 models, and 28,516 applications across Hugging Face and GitHub. We find that an astonishing
96.5\% of datasets and 95.8\% of models lack the required license text, only 2.3\% of datasets and 3.2\% of models
satisfy both license text and copyright requirements, and even when upstream artifacts provide complete licensing
evidence, attribution rarely propagates downstream: only 27.59\% of models preserve compliant dataset notices and
only 5.75\% of applications preserve compliant model notices (with just 6.38\% preserving any linked upstream notice).
Practitioners cannot assume permissive labels confer the rights they claim: license files and notices, not metadata,
are the source of legal truth. To support future research, we release our full audit dataset and reproducible pipeline.

\end{abstract}

\maketitle
\section{Introduction}
\label{sec:introduction}

The rapid proliferation of modern AI has been fueled by a culture of openness. Permissive licenses, especially MIT, Apache-2.0, and BSD-3-Clause, are the ecosystem's default markers of openness~\cite{stalnaker2025}. For example, 53.5\% of Hugging Face datasets are labelled as MIT or Apache-2.0~\cite{stalnaker2025}. By signaling that artifacts can be reused, modified, and integrated into products, these licenses enable an \textit{AI supply chain}: training data is transformed into pre-trained models, which are then adapted through fine-tuning using additional licensed datasets and artifacts into specialized derivatives, and ultimately integrated as dependencies into downstream software applications~\cite{jiang2024}. Licenses are the mechanism meant to make this reuse lawful: they grant rights to reuse and redistribute an artifact, conditioned on specific obligations.

Permissive does not mean unconditional. Permissive licenses grant broad reuse rights, but only if basic
notice-preservation conditions are met, including providing the full license text and preserving copyright and
attribution notices~\cite{mit_license, apache_license, bsd_3clause}. These elements must exist in the repository to be preserved downstream.
If an AI artifact is labeled ``MIT'' or ``Apache-2.0'' only as metadata, but the repository omits the license text
and associated notices, the scope of the grant becomes legally ambiguous because the operative permission and its
conditions live in the license text~\cite{mit_license, apache_license, bsd_3clause}. We refer to the license text and
notices collectively as the \emph{compliance payload}. When this payload is missing or not preserved, downstream reuse
may fall outside the scope of the license, effectively reverting the work to default copyright, meaning no one except
the author has permission to use, distribute, or modify it~\cite{mit_license, apache_license, bsd_3clause,
meeker2023open, jacobsen_v_katzer_2008, laurent2004understanding}.

We call this gap between permissive \emph{labels} and the in-repository documentation needed to substantiate them as
\emph{permissive washing}. It creates a false signal of legal safety: an artifact appears permissive, yet the
documentation required to rely on the declared grant may be missing, making downstream reuse legally uncertain and
potentially outside the scope of the license.

To assess how widespread the risk of \textit{permissive washing} is in the AI supply chain, we construct a provenance graph of 124,278 dataset $\rightarrow$ model $\rightarrow$ application AI supply chains spanning 3,338 datasets, 6,664 models, and 28,516 GitHub applications (Section~\ref{sec:approach}). Existing audits largely stop at \emph{declared} license metadata on individual platforms~\cite{stalnaker2025,longpre2024}.
What remains untested at scale is whether the in-repository compliance payload that makes permissive reuse legally
safe is actually present and preserved end-to-end. To measure this, we construct a provenance graph and isolate
28{,}724 dataset$\rightarrow$model$\rightarrow$application AI supply chains where all three artifacts are labeled under
dominant permissive licenses (MIT, Apache-2.0, BSD-3-Clause)~\cite{stalnaker2025,meeker2020open_source_business,pli2022_beyond_open_data}.
We then perform two audits: an \textbf{integrity audit} that checks whether license text and rights-holder copyright
notices are present, and an \textbf{attribution audit} that traces whether upstream copyright notices are preserved
across transitions. Finally, we analyze the \emph{compliance payload gap} to characterize how often upstream artifacts
omit the files that permissive licensing assumes (Section~\ref{subsec:license_erosion_discussion}).\\

\noindent\textbf{Our integrity audit reveals:}
{\setlength{\leftmargini}{1em}
\begin{itemize}
    \item \textbf{Missing license text:} Only 3.5\% of permissively-labeled datasets and 4.2\% of permissively-labeled models include the full license text, i.e., 96.5\% and 95.8\% are missing it (Table~\ref{tab:license_integrity}). When license text is provided, established practice is to include it in a dedicated root \texttt{LICENSE} file~\cite{github_licensing_2024, homeoffice_open_source_licensing_2025}. \textit{Implication:} Permissive labels frequently function as metadata signals rather than actionable license
grants, leaving downstream reuse legally uncertain when the license text is absent.
    \item \textbf{Platform divide:} GitHub applications achieve 74.2\% full compliance, while upstream Hugging Face artifacts achieve only 2.3\% (datasets) and 3.2\% (models) (Table~\ref{tab:license_integrity}); among major Hugging Face organizations in our corpus, only 1.5\% of artifacts satisfy both license text and copyright requirements (Table~\ref{tab:org_compliance}). \textit{Implication:} The AI ecosystem has adopted open-source labels without the compliance practices that accompany them in traditional software.
\end{itemize}
}

\noindent\textbf{Our attribution audit reveals:}
{\setlength{\leftmargini}{1em}
\begin{itemize}
    \item \textbf{Attribution loss:} Even when upstream datasets and models include both license text and a rights-holder
    notice, attribution rarely propagates: only 27.59\% of trained models retain the dataset notice, and only
    5.75\% of applications retain the model notice (Table~\ref{tab:attribution_preservation}).
    \textit{Implication:} users who train on these datasets or ship applications with these models can be \emph{unable to
    satisfy} notice-preservation conditions, putting their use and distribution potentially outside the scope of the
    declared permissive license.
    \item \textbf{Application gap:} Among applications that depend on at least one such dataset or model, only
    \textbf{6.38\%} preserve any linked upstream rights-holder notice (Table~\ref{tab:attribution_preservation}).
    \textit{Implication:} most applications provide no auditable attribution trail, so developers cannot demonstrate
    compliance with permissive attribution obligations when distributing downstream.
\end{itemize}

Taken together, these faults are not merely clerical: they determine where legal risk accumulates in the AI supply chain. For dataset $\rightarrow$ model transitions, developers often look to fair use, but fair use is decided case-by-case: recent cases illustrate divergent outcomes (e.g., Kadrey v. Meta (2025) vs. Thomson Reuters v. Ross Intelligence (2025))~\cite{kadrey2025meta,ThomsonReuters_v_Ross_2025}. Both litigation and the U.S. Copyright Office's recent analysis emphasize whether the new work competes with the original and whether training retains copies of copyrighted material~\cite{bartz_v_anthropic_2024,uscopyright2025part3}. For model $\rightarrow$ application distribution, the law is more settled: models function as software dependencies~\cite{jiang2024}, and open-source conditions have been treated as enforceable copyright conditions~\cite{jacobsen_v_katzer_2008,SFC_v_BestBuy_2009,Entouvert_v_Orange_2024}. In both cases, compliance payloads are the dividing line: with them, developers can point to verifiable permission; without them, they face either uncertain fair use defenses or direct infringement risk. Our audit shows that only 2.3\% of datasets and 3.2\% of models provide them (Table~\ref{tab:license_integrity}).

To understand why fewer than 3.2\% of upstream artifacts achieve full compliance, we examine a structural fault: the compliance payload gap, where upstream artifacts lack the required files for redistribution. This positions our core contribution: permissive labels are an unreliable proxy for legal usability unless the required license text and attribution payloads are actually present and preserved.

\noindent\textbf{Our Contributions.} 
{\setlength{\leftmargini}{1em}
\begin{itemize}
    \item \textbf{Characterization of permissive washing:} We define and quantify the practice of using permissive labels to mask technically undocumented (and potentially unlicensed) AI artifacts.
    \item \textbf{Large-scale provenance audit:} We provide an end-to-end study of 124,278 dataset $\rightarrow$ model $\rightarrow$ application AI supply chains, including 28,724 permissive lineages.
    \item \textbf{Quantification of compliance fault:} We measure the absence of license text, copyright notices, and attribution at each AI supply chain transition.
    \item \textbf{Open compliance resources:} We release a replication package with our audit data and reproducible pipeline to enable the community to verify and repair AI supply chain license integrity.
\end{itemize}
% \gopi{Also consider saying right away or first in the intro our methodology is simple and deterministic and it already reveals so many problems (kinda defends against you did something simple critique) and mention it in the conclusion}
\section{Related Work}
\label{sec:related}
\begin{figure*}[t]
    \centering
    \includegraphics[width=0.9\textwidth]{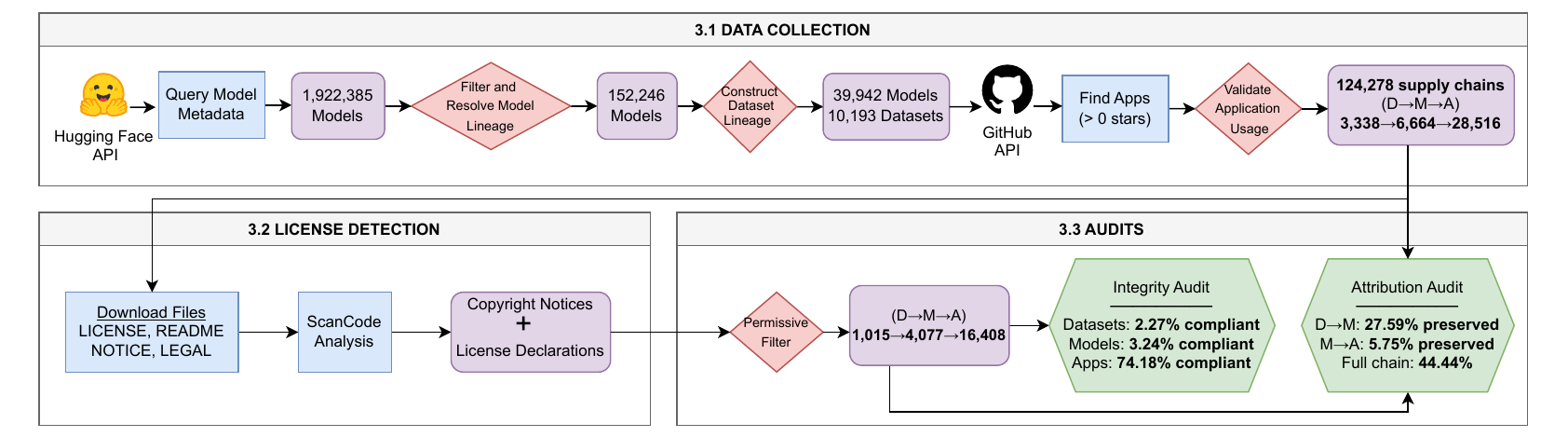}
    \caption{An overview of our three-stage methodology: data collection, license detection, and compliance auditing.}
    \label{fig:overview}
\end{figure*}
% \gopi{This can look smaller..Lots of white space even moving the output box between 3.1 to 3.2 to the side will give you more space. Bold final numbers }
% ● DO provide a brief survey of the relevant related work
% – Make sure to cite relevant work from potential reviewers
% – Make sure to check previous conference proceedings or journal volumes
% ● DO group the discussed related work
% – DO NOT just give a laundry list of papers
% – DO NOT provide irrelevant points about the prior work (e.g., if you do not
% care about the studied systems do not mention them)
% ● DO emphasize why your work is different/important
\textbf{Empirical audits of AI supply chains and licensing.}
Recent work has begun mapping licensing and documentation in AI artifact ecosystems. Stalnaker et al.~\cite{stalnaker2025} conducted the largest analysis of Hugging Face licensing to date, documenting license distributions and metadata inconsistencies between datasets and trained models. Pepe et al.~\cite{Pepe2024HFDocumentation} studied how Hugging Face models document training data, bias considerations, and licenses, finding substantial gaps in model card completeness. At the model$\rightarrow$application boundary, Jiang et al.~\cite{jiang2024} introduced PeaTMOSS, linking pre-trained models to downstream GitHub repositories and showing that developers integrate models as software dependencies.

Complementary audits examine whether licensing labels faithfully reflect upstream sources. The Data Provenance Initiative (Longpre et al.)~\cite{longpre2024,Longpre2025Multimodal} reported that many datasets lack explicit license documentation and that permissive declarations can be inconsistent with underlying sources, a phenomenon often discussed as rights or license laundering in practice. Kim et al.~\cite{Kim2025NEXUS} argue that compliance requires lifecycle tracing because rights and obligations can be lost across redistribution. Platform design further shapes these norms; Gorwa and Veale~\cite{Gorwa2024ModelMarketplaces} analyze governance across model marketplaces (including Hugging Face and GitHub) and highlight how platform choices influence ecosystem practices.

\vspace{0.2em}
\noindent\textit{\textbf{Gap and contribution.}}
Prior studies either analyze stages in isolation or treat declared metadata as ground truth. Stalnaker et al.\ examine dataset$\rightarrow$model patterns but stop at the model stage. Jiang et al.\ trace model$\rightarrow$application dependencies but use the model's declared license as the downstream license label. Longpre et al.\ quantify missing or inconsistent license \emph{labels}, but do not test whether the \emph{compliance payloads} required for redistribution are present. Longpre et al.\ also use the term \emph{license laundering} to describe cases where platform-hosted datasets are relabeled into more permissive use categories than their creators intended. Our concept of permissive washing is complementary: even when an artifact is genuinely permissive, the compliance payload required to exercise that permission is absent. We trace complete dataset$\rightarrow$model$\rightarrow$application lineages and audit file-level payloads at each stage. Furthermore, we quantify whether copyright attribution required by permissive licenses propagates across transitions.

\textbf{Documentation Frameworks and Compliance Standards:}
Software supply chains commonly operationalize licensing and provenance through Software Bills of Materials (SBOMs). They do so by providing a documentation of all the components, packages, thier licenses and provenance used in the software. SPDX is a widely used SBOM standard that formalizes license and provenance reporting as payload-grounded documentation intended to persist through redistribution. This standard is being extended with AI-focused profiles (e.g., AI BOM concepts) to represent datasets, models, and their relationships~\cite{spdxspec,spdx_aibom}. In parallel, the AI community has proposed standardized frameworks for documenting dataset and model provenance. Gebru et al.~\cite{gebru2021datasheets} introduced Datasheets for Datasets, a protocol for disclosing composition, collection process, intended uses, and licensing. Mitchell et al.~\cite{mitchell2019model} proposed Model Cards to document model performance, limitations, and ethical considerations. 

Beyond documentation, researchers have developed compliance-focused tools and standards. Contractor et al.~\cite{Contractor2022} created the Responsible AI License (RAIL) framework, extending traditional open source licenses with behavioral use restrictions prohibiting applications such as surveillance or misinformation generation. CycloneDX ML-BOM~\cite{CycloneDXMLBOM} extends SBOM frameworks to AI models, treating them as components with provenance and license obligations. Duan et al.~\cite{duan2024modelgo} introduced ModelGo, a license compliance analyzer that detects conflicts in AI workflows by examining dependency graphs.

\vspace{0.2em} 
\noindent\textit{\textbf{Gap and contribution.}}
These frameworks specify \emph{what} to record and \emph{how} to reason about conflicts, but they assume that license information exists and is reliable. ModelGo checks conflicts between declared licenses, CycloneDX specifies metadata to populate, and RAIL defines terms to propagate. We test this prerequisite directly by verifying whether license texts, notices, and attribution evidence are present in distributed AI artifacts. This shifts the emphasis from conflict detection to \emph{existence verification}, which is necessary when permissive washing strips the compliance payloads that tooling expects to analyze.

\textbf{License compliance in open source software (OSS).}
License compliance has been extensively studied in traditional OSS ecosystems. Wu et al.~\cite{Wu2024OSLicense} report license incompleteness rates of 14.91\% in Maven, 7.3\% in NPM, and 13.43\% in PyPI. Riehle et al.~\cite{Riehle2022GitHub} find that 34\% of GitHub repositories lack a \texttt{LICENSE} file and 50\% do not fully declare licensing terms. Jahanshahi et al.~\cite{Jahanshahi2024OSS} applied license detection tools including ScanCode at billion-file scale within World of Code, demonstrating scalable methods for OSS license identification.

\vspace{0.2em}
\noindent\textit{\textbf{Gap and contribution.}}
These OSS studies provide a baseline for mature ecosystems where licensing is anchored in file-based conventions and routinely consumed by tooling. In contrast, we find substantially higher absence of compliance payloads in AI hubs (exceeding 90\%), indicating a regression from payload-grounded practices toward metadata-only labels. Methodologically, OSS audits typically operate within a single repository or package ecosystem, whereas AI compliance requires cross-platform linkage and artifact validation across dataset$\rightarrow$model$\rightarrow$application transitions, which is the focus of our pipeline and measurements.

\section{Methodology}
\label{sec:approach}
% ● DO provide enough detail for replication (parameters, thresholds,
% subject selection, etc.)
% ● DO provide an overview figure of your steps
% – Usually a data flow diagram
% – Describe every element in the figure in the text
% – Component titles must use exact same wording as subsection headings
% ● DO choose a realistic baseline to compare your approach against
% ● DO NOT mix conceptual approach and implementation

Our methodology for an end-to-end audit of the open source AI supply chain comprises three stages, as illustrated in Figure~\ref{fig:overview}. This pipeline is deliberately simple and deterministic, relying on proven pattern-based file retrieval and established license-detection tooling with clearly defined thresholds. The \textbf{first stage is data collection}, where we construct complete AI supply chains linking datasets and models from Hugging Face to downstream applications on GitHub. The \textbf{second stage is license detection}, where we extract and identify license declarations and copyright information. The \textbf{third stage is auditing}, where we evaluate payload compliance integrity and measure attribution preservation across permissive AI supply chains. In our auditing stage, we operationalize ``permissive'' via the license labels identified, not by inferring author intent. 

% \gopi{Say we purposely chose a deterministic simple pipeline but we do it carefully}

\subsection{Stage 1: Data Collection}
\label{subsec:data_collection}
Our goal is to construct complete AI supply chains linking training datasets and upstream models on Hugging Face to downstream applications on GitHub. We therefore (i) identify models and their ancestry, (ii) resolve the datasets used to train those models, and (iii) validate downstream model usage in GitHub projects, filtering out incomplete and toy artifacts.

\noindent \textbf{Step 1 (Querying Model Metadata).} We query all model metadata from Hugging Face, the dominant open-source hub for AI models and datasets~\cite{stalnaker2025}, yielding 1,922,385 models as of August 08, 2025.

\noindent \textbf{Step 2 (Constructing model lineage).} We filter to retain models with at least one ``like'', yielding 148,729 models. This threshold excludes projects without community engagement while preserving a large-scale corpus for analysis. We use the number of likes as our popularity metric because recent work shows that this is a strong proxy for community adoption and, unlike download counts, is consistently available via the Hugging Face API~\cite{kadasi2025modelhubsbeyondanalyzing}. We then resolve model lineage by following \texttt{base\_model} references to identify ancestor models, adding them regardless of their number of likes. This upstream resolution yields a total of 152,246 models.

\noindent \textbf{Step 3 (Constructing dataset lineage).} A model is said to have dataset lineage if its \texttt{datasets} field lists at least one training dataset (or if any ancestor model does). Since models without dataset lineage cannot form complete AI supply chains we have to remove these models. We extract dataset references from all models in our corpus (both filtered models and their ancestors) using the \texttt{datasets} metadata field provided in the model card. Since dataset identifiers are often incomplete (e.g., missing organization prefixes), we resolve them by querying Hugging Face to obtain fully qualified identifiers, successfully resolving 10,193 of 12,514 unique references. This yields 39,942 unique models linked to 10,193 unique datasets. 

\noindent \textbf{Step 4 (Validating application usage and complete AI supply chains).} We identify downstream applications via the GitHub Code Search API, querying for each model using its fully qualified identifier (e.g., \texttt{microsoft/codebert-base}), searching across all file types. We then filter the results to Python source files, as Python dominates ML model integrations (see Appendix~\ref{app:file_types} for file type distribution), yielding 721,799 unique .py files across 194,674 repositories. To confirm actual model usage rather than incidental string matches, we subsequently parse each Python file into an Abstract Syntax Tree~(AST) using Python's native AST library and validate against a curated set of 8,968 model-loading signatures, following prior work on model-usage detection~\cite{jiang2024}. This validation confirms that 167,959 applications contain actual model usage, which we filter further to exclude toy projects  without any stars~\cite{Dabic2021}, retaining 66,819 applications. Stars on GitHub serve a similar function to likes on Hugging Face as a proxy for community engagement. Finally, we remove models and datasets with no downstream applications, retaining only complete AI supply chains.

\noindent \textbf{Output.} Our final corpus comprises 124,278 validated AI supply chains spanning 3,338 datasets, 6,664 models, and 28,516 applications.

\subsection{Stage 2: License Detection}
\label{subsec:license_detection}

\noindent \textbf{Overview.} We extract license labels from both platform metadata and repository contents by retrieving likely license-relevant files via filename patterns (Step~2) then analyzing them with ScanCode Toolkit (Step~3).

\textbf{Step 1 (Collecting platform metadata).} We collect license metadata from both platforms in our AI supply chain. For Hugging Face artifacts, we extract the \texttt{license} label from model and dataset metadata, which Hugging Face renders directly from the model card's YAML header~\cite{huggingface_hub_model_cards_guide}. For GitHub applications, we derive license information directly from the file-level analysis (Step 3), which provides broader coverage than GitHub's built-in license detection (limited to 49 known licenses)~\cite{github_licensing_2024}.

\textbf{Step 2 (Download Files).} Beyond metadata, we extract license text directly from AI artifact files. For permissive licenses (MIT, Apache-2.0, BSD-3-Clause), redistribution requires preserving license text and notices/attribution information~\cite{mit_license, apache_license, bsd_3clause}. By convention, these compliance payloads are typically placed in top-level files such as \texttt{LICENSE}, \texttt{NOTICE}, and \texttt{COPYRIGHT}, and sometimes \texttt{README}. We therefore retrieve four compliance-relevant file classes (license/notice/copyright artifacts and README) using extensive case-insensitive regex patterns over filenames and directories, following prior conventions~\citet{Zacchiroli_2022} (we provide the patterns in Appendix~\ref{app:file_patterns}). Since license content can appear in non-standard locations, we validate the \emph{retrieval coverage} of this pattern-based stage against full-repository ScanCode scans on a representative sample (Appendix/Table~\ref{tab:license_coverage_validation}).

\textbf{Step 3 (ScanCode analysis and extraction).} We analyze the files downloaded in Step 2 using the ScanCode Toolkit~\cite{ScanCodeToolkit}, which uses a search index of over 32,000 license texts and variants to detect licenses across software, data, and documentation. The toolkit employs token-based matching with multiple fallback strategies to handle variations in license text formatting and completeness. We extract two outputs from ScanCode: (i) license declarations (whether the declared license text is present) and (ii) copyright notices (statements identifying copyrights holders).

\noindent \textbf{Output.} For each AI supply chain in our corpus, we extract license declarations and copyright notices for every AI artifact in the supply chain (datasets, models, and applications), using ScanCode outputs for authoritative file-level payload information.

\subsection{Stage 3: Auditing} 
\label{subsec:audits}

As described in Section~\ref{sec:introduction}, permissive licenses impose concrete redistribution obligations,
including preserving license text, copyright notices, and attribution. We operationalize these obligations
through two audits: (i) the License Integrity Audit and (ii) the License Attribution Audit. The audits share
the same end-to-end dataset$\rightarrow$model$\rightarrow$application corpus (Section~\ref{subsec:data_collection})
and the extracted licensing signals and files (Section~\ref{subsec:license_detection}), but apply audit-specific
filtering as needed.

\subsubsection{License Integrity Audit}
\label{subsubsec:integrity_method}

For our license integrity audit, we assess whether permissively labeled AI artifacts
satisfy basic, file-level license-documentation conditions needed to credibly support their permissive label:
(i) the license text and (ii) a copyright notice/statement.

\noindent \textbf{Approach.} Using the license metadata extracted
in Section~\ref{subsec:license_detection}, we restrict analysis to fully permissive dataset$\rightarrow$model$\rightarrow$application
AI supply chains, where the dataset, model, and application are each labeled MIT, Apache-2.0, or BSD-3-Clause. This
filter yields 28{,}724 AI supply chains. For the integrity audit, we evaluate all permissively labeled AI artifacts
within these chains: 1{,}015 datasets, 4{,}077 models, and 16{,}408 applications.

For each dataset, model, and application in this permissive subset, we extract two elements from the ScanCode
analysis (Section~\ref{subsec:license_detection}): (1) license declarations (license text coverage), and (2)
copyright statements. We treat license text as present if ScanCode detects the full license text at $\geq$ 90\%
coverage, following established similarity thresholds in software engineering research~\cite{Golubev2020}. We
treat copyright as present if ScanCode detects at least one copyright statement (e.g., ``Copyright 2024
Organization Name'').

We define an AI artifact as \textit{compliant} only if both elements are present, because permissive reuse is
conditioned on preserving license text and notices; when either is missing, the grant becomes ambiguous and
downstream reuse may fall outside the scope of the license (effectively reverting to default copyright for those
uses)~\cite{laurent2004understanding}. We evaluate compliance separately for datasets, models, and applications
to localize integrity failures along the AI supply chain. Finally, to assess whether integrity failures persist among
widely adopted publishers, we examine verified Hugging Face organizations whose permissively labeled AI artifacts
appear in our corpus. Ranking these organizations by follower count and selecting the top 15 yields 462 artifacts.

\noindent \textbf{Outcome.} Artifact-level integrity labels (compliant/non-compliant) for datasets, models, and
applications, plus prominence-stratified compliance summaries for the top 15 verified organizations.

\subsubsection{License Attribution Audit}
\label{subsubsec:attribution_method}
For the license attribution audit, we assess whether rights-holder attribution evidence is preserved when AI
artifacts are used downstream. We operationalize attribution evidence as file-level copyright notices, since they
identify rights holders and provide a concrete notice-preservation target in permissive licensing
contexts~\cite{Xu2025}.

\noindent \textbf{Approach.} We build this audit on the license integrity audit
(Section~\ref{subsubsec:integrity_method}). We first define \textit{compliant} datasets and models as those that
contain both (i) the full license text and (ii) at least one valid rights-holder copyright notice, as detected by
ScanCode. We then test whether the upstream copyright notice text appears downstream. In all attribution slices
below, downstream artifacts are drawn from the full AI supply-chain corpus and are not restricted to the permissive
subset. We use CD to denote compliant datasets, CM to denote compliant models, M to denote models, and A to denote
applications.

\textbf{Normalization and matching.} For each compliant upstream artifact, we extract its copyright notice text
from ScanCode as the authoritative attribution string. To account for formatting variation, we normalize both
upstream notices and downstream repository text by removing non-alphanumeric characters and converting to
lowercase~\cite{Choi2015}. We treat attribution as preserved if the normalized upstream notice appears as a
substring in the normalized downstream text, which exactly detects literal overlaps~\cite{Gipp2011}.

\textbf{Slice S1 (CD$\rightarrow$M).} \emph{Compliant datasets to models.} For each compliant dataset, we enumerate
all models linked to it in our supply-chain corpus and test whether the dataset notice is preserved in each linked
model.

\textbf{Slice S2 (CM$\rightarrow$A).} \emph{Compliant models to applications.} For each compliant model, we
enumerate all applications linked to it in our supply-chain corpus and test whether the model notice is preserved
in each linked application.

\textbf{Slice S3 ((CD$\cup$CM)$\rightarrow$A).} \emph{Any compliant upstream to applications.} We consider all
applications that are linked to at least one compliant dataset or at least one compliant model. For each such
application, we test whether it preserves any linked compliant dataset notice, any linked compliant model notice,
or both.

\textbf{Slice S4 (CD$\rightarrow$CM$\rightarrow$A).} \emph{Joint compliant dataset and compliant model to
applications.} We identify applications that lie on at least one dataset$\rightarrow$model$\rightarrow$application
path where both the dataset and the model are compliant. For these applications, we test joint preservation by
measuring whether the application preserves both notices simultaneously.

\noindent \textbf{Outcome.} Preservation indicators for each slice (S1 to S4), enabling transition-level and
application-level summaries of dataset attribution, model attribution, and joint attribution preservation.

% \section{Results}
% \label{sec:results}

% \input{sections/exploratory}
\section{License Integrity Audit}
\label{sec:integrity_results}
\begin{table}[t]
\centering
\small
\caption{License integrity of permissively-labeled AI artifacts (MIT, Apache-2.0, BSD-3-Clause). Has License Text: full text detected ($\ge$90\% match). Has Copyright: valid notice identifying a rights holder. Fully Compliant: both present.}
\label{tab:license_integrity}
\begin{tabular}{@{}lrrrr@{}}
\toprule
\textbf{Artifact} & \textbf{Total} & \textbf{\shortstack{Has License\\Text}} & \textbf{\shortstack{Has\\Copyright}} & \textbf{\shortstack{Fully\\Compliant}} \\
\midrule
Datasets     & 1,015    & 36 (3.5\%)      & 30 (3.0\%)       & 23 (2.3\%)      \\
Models       & 4,077    & 171 (4.2\%)     & 197 (4.8\%)      & 132 (3.2\%)     \\
Apps         & 16,408   & 15,081 (91.9\%) & 12,537 (76.4\%)  & 12,172 (74.2\%) \\
\bottomrule
\end{tabular}
\end{table}

% \findings

We evaluate whether permissively-labeled AI artifacts include the required compliance payload, as described in Section~\ref{subsubsec:integrity_method}. Our permissive subset comprises 1,015 datasets, 4,077 models, and 16,408 applications.

\textbf{License text is largely missing on Hugging Face, even when AI artifacts are labeled permissive.} In our permissive subset, only 3.5\% of datasets and 4.2\% of models include a complete license text, compared to 91.9\% of applications (Table~\ref{tab:license_integrity}). This is not confined to obscure artifacts: widely-used AI artifacts such as Anthropic/hh-rlhf (MIT, 1,391 likes) and sentence-transformers/all-MiniLM-L6-v2 (Apache-2.0, 3,737 likes) both lack the full license text needed to confidently rely on the declared permissive terms in practice. This risk compounds downstream: sentence-transformers/all-MiniLM-L6-v2 enables 907 downstream applications (Section~\ref{subsec:license_erosion_discussion}),which cannot preserve the upstream license text as required by notice-preservation conditions
because it is absent at the source.

% \hao{Fix the precision of the numbers, they are different from the ones in the table, e.g., 2.19\% vs. 2.2\%}
% \hao{report the exact percentage}

\textbf{Rights-holder copyright notices are also frequently missing, leaving even ``compliant-looking'' AI artifacts
legally ambiguous.} Only 3.0\% of datasets and 4.8\% of models include a valid copyright notice identifying a rights
holder, compared to 76.4\% of applications (Table~\ref{tab:license_integrity}). Notably, even AI artifacts that
include license text often omit a rights-holder statement: \texttt{OpenAssistant/oasst1} (Apache-2.0, 1{,}422 likes)
and \texttt{google-bert/bert-base-uncased} (Apache-2.0, 2{,}365 likes) both provide license text but do not declare
a copyright holder. In such cases, downstream users lack a clear, file-level attribution target and may be unable
to satisfy notice-preservation obligations under permissive terms, making compliant reuse harder in
practice.

\textbf{Even major AI organizations are overwhelmingly non-compliant.} Table~\ref{tab:org_compliance} in Appendix~\ref{app:org_compliance} reports compliance for the top 15 verified Hugging Face organizations (by follower count) with permissively-labeled AI artifacts in our supply chain corpus. License text is nearly absent: only Intel (20.0\%), Ai2 (5.1\%), and Microsoft (1.7\%) include any license text, while the remaining 12 organizations show 0\% license text presence. Copyright notices are slightly more common, but still rare: Google leads at 11.2\%, followed by Ai2 (6.3\%) and AI at Meta (4.5\%). Overall, only 7 of 462 AI artifacts (1.5\%) from these organizations satisfy both requirements. If well-resourced organizations with dedicated legal teams cannot achieve compliance for AI artifacts that they actively distribute, the expectation that individual developers will do so is unrealistic.

% \gopi{bold strategically in all summary/takeaway places} \jim{updating this!}
\hypobox{\textbf{Permissive labels are not reliable evidence of reusable AI artifacts under the declared terms.} Only \textbf{2.3\%} of datasets and \textbf{3.2\%} of models are fully compliant (license text and copyright notice present), while \textbf{74.2\%} of applications are. \textit{Implication:} legal risk concentrates upstream, and downstream users inherit that risk because the required documentation does not exist.}

\section{License Attribution Audit}
\label{sec:attribution_results}

\begin{table}[t]
\centering
\small
\caption{Attribution preservation from \textit{compliant} upstream AI artifacts. We trace file-level copyright
notices from compliant datasets and compliant models (defined in Section~\ref{subsubsec:integrity_method}) into
downstream artifacts in the full supply-chain corpus. Rows correspond to audit slices S1 to S4.}
\label{tab:attribution_preservation}
\begin{tabular}{@{}lrrrr@{}}
\toprule
\textbf{Slice} & \textbf{Evaluated} & \textbf{Preserved} & \textbf{Not Preserved} & \textbf{Rate} \\
\midrule
S1 (CD$\rightarrow$M) & 58 & 16 & 42 & 27.59\% \\
S2 (CM$\rightarrow$A) & 2{,}104 & 121 & 1{,}983 & 5.75\% \\
S3 ((CD$\cup$CM)$\rightarrow$A) & 2{,}227 & 142 &  2{,}085 & 6.38\% \\
S4 (CD$\rightarrow$CM$\rightarrow$A) & 27 & 12 & 15 & 44.44\% \\
\bottomrule
\end{tabular}
\end{table}

We evaluate whether attribution evidence from upstream AI artifacts is preserved downstream, as described in
Section~\ref{subsubsec:attribution_method}. This audit is intentionally conservative. Rather than starting from all
permissively labeled artifacts, we start from the small subset with \textit{compliant} licenses in the sense of
Section~\ref{subsubsec:integrity_method}, meaning the repository contains both the full license text and at least one
valid rights-holder copyright notice. As shown in Table~\ref{tab:license_integrity}, such artifacts are rare among
Hugging Face datasets and models, which limits the number of upstream artifacts for which attribution preservation is
even well-defined.

Starting from the compliant upstream datasets and models identified by the integrity audit, we trace their usage in
the full supply-chain corpus and evaluate four attribution slices. We evaluate 58 dataset$\rightarrow$model links
originating from compliant datasets (S1), 2{,}104 model$\rightarrow$application links originating from compliant
models (S2), 2{,}227 applications that are linked to at least one compliant dataset or compliant model (S3), and 12
applications that lie on at least one dataset$\rightarrow$model$\rightarrow$application path where both the dataset
and the model are compliant (S4). Table~\ref{tab:attribution_preservation} summarizes preservation outcomes across
these slices.

\textbf{Even when dataset licensing evidence is complete, attribution is frequently dropped during training.}
Among dataset$\rightarrow$model links originating from compliant datasets, only 16 of 58 (27.59\%) preserve the
dataset's copyright notice in the downstream model (Table~\ref{tab:attribution_preservation}). Put differently, most
derived models omit the dataset rights-holder notice even when it is present and well-formed upstream.

\textbf{Attribution collapses when models are integrated into applications.}
Among model$\rightarrow$application links originating from compliant models, only 121 of 2{,}104 (5.75\%) preserve
the model's copyright notice in the downstream application (Table~\ref{tab:attribution_preservation}). This indicates
that application integration is the dominant point of attribution loss, even in the best-case setting where upstream
models ship complete license text and an explicit rights-holder notice.

\textbf{Across the application layer, preserving any upstream attribution is uncommon.}
Considering applications linked to at least one compliant dataset or at least one compliant model, only 142 of 2{,}227
applications (6.38\%) preserve any linked upstream copyright notice (dataset, model, or both)
(Table~\ref{tab:attribution_preservation}). Most applications therefore provide no file-level attribution target for
the compliant upstream artifacts they depend on.

\textbf{Joint preservation is feasible when both upstream notices exist, but this scenario is rare.}
In the small subset of applications that lie on at least one dataset$\rightarrow$model$\rightarrow$application path
where both the dataset and the model are compliant, 12 of 27 applications (44.44\%) preserve both notices
simultaneously (Table~\ref{tab:attribution_preservation}). This suggests that when both upstream notices exist and are
traceable, dual preservation is achievable, but such jointly compliant paths are uncommon in practice.

\hypobox{\textbf{Attribution does not reliably propagate, even from the rare AI artifacts with compliant licenses.}
Only \textbf{27.59\%} of dataset$\rightarrow$model links preserve dataset notices and only \textbf{5.75\%} of
model$\rightarrow$application links preserve model notices; at the application layer, just \textbf{6.38\%} of
applications preserve any linked upstream notice. \textit{Implication:} attribution obligations are routinely
broken, so downstream users cannot assume they have retained the attribution evidence required to rely on permissive
terms, increasing legal and compliance risk.}

\section{Compliance Payload Gap}
\label{subsec:license_erosion_discussion}

While trying to understand the astonishingly common practice of permissive washing in end-to-end AI supply chains, we found that the root cause is more fundamental: the information needed to even \emph{form} traceable chains is routinely missing. While our audit methodology started off with 1,922,385 Hugging Face models (Stage~1, Step~1), only 136,890 (7.1\%) actually were found to declare training datasets via metadata (Appendix~\ref{app:traceability_artifacts}). After additionally requiring validated downstream usage in GitHub applications and pruning incomplete chains (Section~\ref{subsec:data_collection}), we were left with 124,278 complete dataset $\rightarrow$ model $\rightarrow$ application supply chains spanning 3,338 datasets, 6,664 models, and 28,516 applications. Our main audit then further narrowed this data set to 1,015 datasets, 4,077 models and 16,408 fully permissive end-to-end chains in Stage 3 (integrity and attribution audit). 

This compounding attrition is not just a sampling nuisance: it highlights a structurally thin ``traceable'' slice of the open AI ecosystem, where missing provenance and missing compliance payloads can obscure legal obligations. To interpret what this thin traceable slice implies for compliance, we examine the structural fault underlying these failures.

The supply chains we can audit are bounded by disclosure: only 7.1\% of models (136,890 of 1,922,385) on Hugging Face declare their training datasets via metadata (Appendix~\ref{app:traceability_artifacts}). Fine-tuning lineage is more common (29.6\% declare a base model), but complete provenance remains the exception. As a result, large parts of the ecosystem are effectively out of scope for any end-to-end compliance check because provenance is untraceable.

% \gopi{Need to make this stronger with that screenshot from hugging face that says use a tag and state it in readme} \jim{added screenshot! note - I have an alt version fig3\_alt if the text looks too small}

Even within the traceable slice, we observe a more immediate gap that directly supports our license integrity audit results: permissive labels frequently lack the compliance payloads required for reuse. Permissive licenses require that the full license text accompanies redistributed AI artifacts; standard open-source practice operationalizes this via a dedicated root \texttt{LICENSE} file~\cite{github_licensing_2024, homeoffice_open_source_licensing_2025}. However, this convention is largely absent upstream: 96.5\% of datasets and 93.4\% of models in our corpus lack a \texttt{LICENSE} file, compared to only 39.8\% of applications (Table~\ref{tab:missing_artifacts}, Appendix~\ref{app:traceability_artifacts}). README files show the inverse pattern (common on Hugging Face but rare on GitHub), consistent with workflows that treat license \emph{selection} as card metadata in the README/model card rather than shipping a dedicated \texttt{LICENSE} File~\cite{huggingface_hub_license_2024}, as required, but not enforced, by Hugging Face's license policy  (Figure~\ref{fig:hf_license_docs}). While license information \emph{can} be embedded in README text, our audit shows that upstream AI artifacts rarely include the full license text and required notices in practice, which helps explain the documentation faults we quantify in our integrity audit.

\textbf{This payload gap is not uniform across platforms.} We observe a stark disparity in artifact-level compliance between the software ecosystem and the AI ecosystem: GitHub applications achieve 74.2\% full compliance (license text and required notices present), while Hugging Face datasets and models are at 2.3\% and 3.2\%. This disparity is consistent with the hypothesis that platform affordances shape community compliance norms. GitHub's workflow centers on file trees, where a root \texttt{LICENSE} file is a standard convention. In contrast, Hugging Face prioritizes metadata and model cards (READMEs). Our data shows that 97.0\% of Hugging Face datasets and models store licensing information (declarations/references) in README text, compared to only 3.0\% using dedicated \texttt{LICENSE} files (Table~\ref{tab:license_locations}, Appendix~\ref{app:license_locations}). By treating licensing primarily as metadata rather than file-based compliance payloads, current hub designs can decouple the appearance of openness from the documentation needed to verify it. 
% \gopi{Maybe we move the bit about huggingface license screenshot here and expand a bit}

\begin{figure}[t]
\centering
\includegraphics[width=\columnwidth]{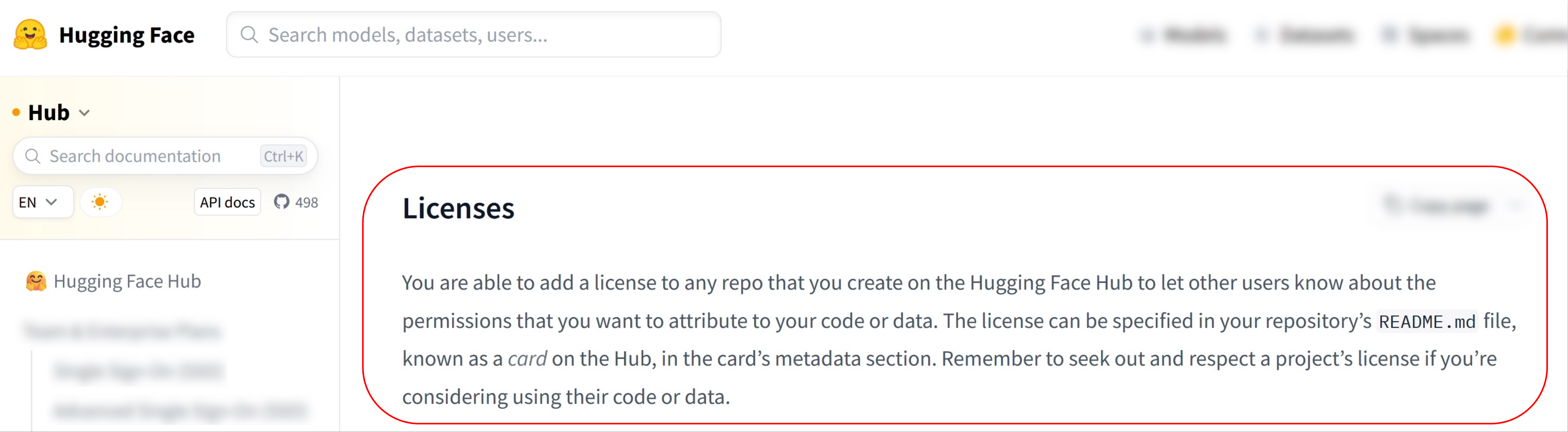}
\caption{Official Hugging Face documentation regarding license specification.}
\label{fig:hf_license_docs}
\end{figure}

\hypobox{\textbf{Takeaway:} Permissive-washing is enabled by a fundamental compliance payload gap: upstream AI artifacts rarely ship the license text and notices that make permissive labels actionable. As a result, permissive labels are an unreliable proxy for legal usability: even ``fully permissive'' chains frequently fail basic documentation and attribution requirements.}

\section{Discussion and Limitations}
\label{sec:discussion}
\subsection{Legal Perspectives on Risk Exposure from Permissive Washing}

At its core, \textbf{permissive washing} in AI supply chains preserves the label of openness while removing the compliance payloads (files and notices) that enable permissions and obligations to be verified and propagated. While our study documents, for the first time at scale, the extent of permissive washing, we are not lawyers and cannot provide legal advice. We do not determine whether a specific dataset license binds a specific model. However, using the legal concept of risk exposure, we can relate our empirical findings to prior and current legal cases and their implications. Across the legal perspectives we examine below, the risk mechanism is consistent: missing compliance payloads undermine the verifiability of permissions at every stage of the supply chain.

\subsubsection{Training legality risk: Permissive washing amplifies uncertainty}
At the ``input'' stage, the legality of training on copyrighted data remains actively litigated and fact-sensitive. In \emph{Thomson Reuters v.\ Ross Intelligence}, the court rejected Ross's fair use defense on summary judgment in a setting where the product arguably competed in an informational market relevant to the training content~\cite{reuters_v_ross_2025}. In \emph{Kadrey v.\ Meta}, the court granted Meta partial summary judgment on fair use on the summary-judgment record before it, while other theories (including an alleged distribution theory tied to torrenting) continued separately~\cite{kadrey_v_meta_2025}. International divergence adds complexity; UK litigation in \emph{Getty Images v.\ Stability AI} has proceeded under different doctrinal and territorial constraints, including narrowing of issues and a UK High Court judgment addressing (UK-law) claims distinct from U.S.\ fair use framing~\cite{Getty_Stability_UK_Judgment_2025}. Scholarship similarly diverges~\cite{henderson2023foundation,charlesworth2025illusory,alhadeff2024limits}.
\\

\noindent\textbf{Risk takeaway:} Fair use in U.S. litigation is a legal defense that places the burden of proof on the defendant, and is often fact-intensive and costly to establish. Payload-backed licensing reduces verification cost and variance; however, our integrity audit found that \textbf{96.5\% of datasets labeled as permissive lack license text} (Section~\ref{sec:integrity_results}, Table~\ref{tab:license_integrity}). By stripping these payloads from their artifacts, permissive washing pushes developers away from verifiable licensing and toward uncertain defenses and incomplete provenance.

\subsubsection{Weights/derivation risk: The payload gap severs the chain of rights}
A second uncertainty concerns whether model weights are ``copies'' or ``derivative works.'' Franceschelli et al.\ connect ``training-as-compression'' to memorization and legal debates about representation~\cite{franceschelli2024compression}. If stricter legal views emerge that treat trained model weights as containing meaningful traces of their training data, the case for preserving and auditing upstream permissions across the chain becomes stronger.
\\
\noindent\textbf{Risk takeaway:} Under stricter derivation views, attribution evidence does not reliably propagate even from the rare upstream artifacts with complete permissive documentation. In our attribution audit, only 27.59\% of dataset→model links preserve dataset notices, and only 5.75\% of model-application links preserve model notices. This indicates a systemic inability to demonstrate that attribution conditions are satisfied downstream, increasing the risk of challenges even when AI artifacts are labeled permissive.

\subsubsection{Redistribution risk: Permissive washing presents a compliance failure}
Regardless of how training disputes resolve, the \emph{redistribution} of models and applications is governed by their licenses. Permissive licenses (MIT, Apache 2.0) commonly condition redistribution on preserving license text and notices; failing to do so is a compliance fault, not a fair use question. Courts have treated open-source license conditions as enforceable copyright conditions (e.g., \emph{Jacobsen v.\ Katzer})~\cite{jacobsen_v_katzer_2008}. Enforcement actions have also targeted distributors/retailers for distributing out-of-compliance software~\cite{SFC_v_BestBuy_2009}, and a French appellate court awarded substantial damages for GPL violations in \emph{Entr'Ouvert v.\ Orange} (2024)~\cite{entrouvert_v_orange_2024}.
\\
\noindent\textbf{Risk takeaway:} Permissive washing creates immediate exposure by decoupling permissive labels from the compliance payload (LICENSE text, notices). With \textbf{93.4\% of model repositories missing license files}, downstream users cannot verify what rights they inherit; they can unknowingly redistribute ``MIT-labeled'' but compliance-incomplete AI artifacts.

\subsubsection{Governance fragility and lifecycle tracing: Weak labels in, data problems out}
The ecosystem increasingly substitutes durable compliance payloads (LICENSE/NOTICE files) with metadata labels and bespoke ``Terms of Use.'' Henderson and Lemley argue that many AI terms-of-use restrictions are legally fragile compared to conditions embedded in the license text itself~\cite{henderson2024mirage}. In parallel, Kim et al.\ argue that one ``cannot trust the licenses you see'' and that compliance requires massive-scale lifecycle tracing to recover lost license and attribution information~\cite{Kim2025NEXUS}; our measurements support that premise because the payload gap removes the audit trail downstream users need.
\\
\noindent\textbf{Risk takeaway:} Permissive washing accelerates drift toward weaker, platform-specific compliance: \textbf{only 3.2\% of models are fully compliant (license text and required notices present)}. These obligations may not travel with artifacts or hold up under scrutiny, and the shift turns compliance from verifying attached documentation into reconstructing obligations from partial compliance, which is precisely the large-scale tracing problem our results motivate (Sections~\ref{sec:integrity_results} and~\ref{sec:attribution_results}).

\subsection{Limitations}
\label{subsec:limitations}

\noindent\textbf{License scope and representativeness.}
We restricted our audit to MIT, Apache-2.0, and BSD-3-Clause. These are widely used permissive licenses and represent a practical baseline for dependency integration in software ecosystems~\cite{meeker2020open_source_business}. We focused on them because they share a standardized structure (broad rights conditioned on notice preservation). We excluded instruments such as CC0 and other Creative Commons variants because they follow distinct legal mechanics and reuse norms (often used for content/data rather than software dependencies)~\cite{meeker2020open_source_business,pli2022_beyond_open_data}. Extending coverage to additional license families may reveal different compliance patterns, which we leave for future work.

\noindent\textbf{The ``traceable slice'' bias.}
Our end-to-end analysis is bounded by the availability of lineage metadata. As noted by Pepe et al.~\cite{Pepe2024HFDocumentation}, only $\approx$14\% of Hugging Face models declare training data in their metadata. Consequently, our chain-level results apply only to the traceable slice of the ecosystem. The direction of bias is ambiguous (e.g., disclosure may correlate with stronger documentation practices), so we avoid extrapolating our end-to-end attribution rate (Section~\ref{sec:attribution_results}, Table~\ref{tab:attribution_preservation}).

\noindent\textbf{Metadata dependence (Appearance vs.\ Intent).}
Our selection of ``permissive'' chains relies on platform metadata labels (e.g., filtering for ``MIT''). If a developer incorrectly tags a proprietary model as MIT, it enters our permissive cohort. While this may mischaracterize the uploader's \emph{intent}, it accurately reflects the \emph{appearance} relied upon by the ecosystem. Downstream developers commonly consume AI artifacts based on the label; therefore, measuring discrepancies between the label and artifact-level compliance evidence is central to the permissive washing phenomenon we quantify.

\noindent\textbf{File retrieval recall.}
To scale our audit to 38,518 AI artifacts (3,338 datasets, 6,664 models, and 28,516 applications), we relied on pattern-based file retrieval rather than full repository clones. Our validation shows that selecting files by filename pattern rather than scanning full repositories achieves 93.6\% recall, meaning we may miss license text in non-standard file locations. Therefore, our results represent a lower bound on detected compliance payloads, though the rate of missing compliance payloads ($>96\%$) is too large to be explained by retrieval error alone.

\noindent\textbf{Attribution matching.} Our attribution audit uses substring matching to detect whether upstream copyright notices appear downstream. This approach provides perfect detection for literal text overlaps~\cite{Gipp2011} but may miss paraphrased or reformatted notices. LLM-based semantic matching could improve recall, which we leave for future work.
 \section{Conclusion}
\label{sec:conclusion}

In this work, we defined and quantified \textbf{permissive washing}: the practice of signaling openness via metadata
while omitting the legal compliance payloads needed to operationalize permissive terms in practice. Our audit of
124{,}278 dataset$\rightarrow$model$\rightarrow$application supply chains shows this is not a fringe error but a
default state upstream: only \textbf{2.3\%} of permissively labeled datasets and \textbf{3.2\%} of permissively labeled
models include both full license text and a rights-holder notice (Table~\ref{tab:license_integrity}). Even when such
complete upstream evidence exists, attribution rarely propagates to downstream use: only \textbf{5.75\%} of
model$\rightarrow$application links preserve the model notice, and only \textbf{6.38\%} of applications preserve any
linked upstream notice (Table~\ref{tab:attribution_preservation}). These failures turn verifiable rights into fragile
labels, leaving downstream developers unable to demonstrate notice-preservation compliance despite ``open'' metadata.
We release our provenance graph and audit pipeline to help the community close this payload gap and restore legal
integrity to the AI supply chain.

\bibliographystyle{ACM-Reference-Format}
\bibliography{sample-base}
\appendix
\setcounter{secnumdepth}{2}
\section{License Categorization Framework}\label{app:category}

\begin{table*}[t] 
\centering
\caption{Definitions of License Categories} 
\label{tab:license_categories} 

\begin{tabular}{p{2.5cm}p{10.5cm}>{\raggedright\arraybackslash}p{3.0cm}}
\toprule
\textbf{Category} & \textbf{Description} & \textbf{Change From Supply Chain 2.0 ~\cite{stalnaker2025}} \\
\midrule

CC RESTRICTIVE & Licenses originating from Creative Commons that impose significant restrictions on use, modification, or distribution. This includes non-commercial and no-derivatives clauses, as well as sharealike variants combined with these restrictions (for example, CC BY-NC-SA).  & Originated from CC class \\

COPYLEFT & Strong copyleft licenses requiring derivatives to be licensed under the same or compatible terms. & Originated from Open Source and Data Class\\

ML\_LICENSE &  Machine Learning Licenses often including specific considerations for how providers use models, as well as obligations on downstream applications similar to copyleft.  & Originated from Machine Learning Class\\

SHARE\_ALIKE & Weak copyleft licenses requiring derivatives or modifications shared under the same or compatible terms but potentially allowing linking with differently licensed code (for example, LGPL, MPL, CC BY-SA). & Originated from Open Source and Data Class \\

PERMISSIVE & Minimal restriction licenses allowing use, modification, and distribution (commercial or non-commercial)  whose primary obligation is attribution (for example, CC BY, MIT, Apache-2.0, BSD). & Originated from Open Source and Data Class\\

PUBLIC DOMAIN & No restrictions (for example, CC0, Unlicense). & Originated from Open Source Data Class\\
\bottomrule
\end{tabular}
\end{table*}

\subsection{Overview of Licensing Paradigms}
% \gopi{ I slotted things indiscrimately in appendix, have some explanation for each thing and make this more structured}
License compliance in the AI supply chain requires navigating three distinct licensing paradigms, each with different philosophical foundations and practical implications.

\subsubsection{Open Source Software Licensing}

Traditional open-source licenses, as defined by the Open Source Initiative (OSI) and championed by the Free Software Foundation (FSF), govern software  distribution and modification. These licenses emerged from a software-centric  worldview and establish obligations primarily around source code sharing and derivative work licensing. Key families include:

\begin{itemize}
\item \textbf{Permissive licenses} (MIT, BSD, Apache-2.0): Minimal restrictions, allowing commercial use and relicensing with only attribution requirements
\item \textbf{Copyleft licenses} (GPL, AGPL): Require derivatives be released under the same or compatible terms, ensuring perpetual openness
\item \textbf{Weak copyleft licenses} (LGPL, MPL): Allow linking with proprietary code while maintaining copyleft for modified portions
\end{itemize}

\subsubsection{Creative Commons Licensing}

Creative Commons licenses were designed for creative works (text, images, datasets) rather than executable software. Critically, \textbf{CC licenses are not open source licenses}; they were explicitly created to address gaps in copyright licensing for non-software artifacts. Many CC licenses include restrictions (NonCommercial, NoDerivatives) that violate the Open Source Definition and are incompatible with software distribution norms.

In the AI supply chain, CC licenses predominate for \textit{datasets}, creating a licensing mismatch when these datasets are used to train models that are then integrated into software. Key CC restrictions include:

\begin{itemize}
\item \textbf{BY (Attribution)}: Requires credit to the creator
\item \textbf{SA (ShareAlike)}: Requires derivatives use the same license
\item \textbf{NC (NonCommercial)}: Prohibits commercial use
\item \textbf{ND (NoDerivatives)}: Prohibits modifications
\end{itemize}

These restrictions create novel compliance challenges. For instance, a CC BY-NC model inherits the NonCommercial restriction, making it legally incompatible with use in any commercial application, even if the application's own code is under a permissive license.

\subsubsection{Machine Learning Specific Licensing}

The emergence of foundation models has spawned a new licensing paradigm that combines elements of open source and Creative Commons approaches with novel \textit{use-based restrictions}. These licenses (e.g., OpenRAIL, BigScience RAIL, Llama Community License) are \textbf{not open source licenses} under the OSI definition, as they impose restrictions on how the model can be used rather than just how it can be distributed.

ML-specific licenses introduce obligations unprecedented in traditional software licensing:

\begin{itemize}
\item \textbf{Ethical use restrictions}: "You may not use this model to generate hate speech or misinformation"
\item \textbf{Domain-specific prohibitions}: "You may not use this model for military or surveillance applications"
\item \textbf{Derivative model obligations}: Requirements that fine-tuned models preserve the original use restrictions
\item \textbf{Downstream propagation requirements}: Applications using the model must include the model's license terms, even if the application code uses a different license
\end{itemize}

This creates a \textit{three-layer licensing challenge}: A permissive software application (MIT license) might use a ML license (OpenRAIL-M) that was trained on a NonCommercial dataset (CC BY-NC-SA), creating a stack of incompatible obligations that no single downstream license can satisfy.

\subsection{License Category Definitions}

Table~\ref{tab:license_categories} provides the consolidated definitions for the license categories used throughout our analysis. Our categorization framework adapts the methodology from~\citet{stalnaker2025}, refined using established principles from the Free Software Foundation~\citep{fsf2025} and Creative Commons~\citep{cc2025}.

\textbf{COPYLEFT vs. SHAREALIKE distinction:} We classify licenses as COPYLEFT only if they require the \textit{entire derivative work} to be released under compatible terms (GPL, AGPL). Licenses that permit more granular copyleft (file-level, module-level, or allowing linking with differently-licensed code) are classified as SHAREALIKE (LGPL, MPL, EPL, CC BY-SA).

\textbf{ML LICENSE category:} We group licenses with use-based restrictions into a dedicated ML LICENSE category because these restrictions represent a fundamentally different compliance model than traditional distribution-focused obligations. These licenses cannot be adequately represented by traditional OSI categories.

\textbf{Creative Commons complexity:} CC licenses combine orthogonal restrictions (BY, SA, NC, ND) that we decompose into separate categories. For instance, CC BY-NC-SA is categorized as NC-SA to capture both the NonCommercial restriction and the ShareAlike obligation.

\section{Methodological Validation}\label{app:validation}

\subsection{Validation of AI Artifacts' License Retrieval Coverage}
\label{app:license_retrieval_validation}

  \begin{table}[t]
  \centering
  \sffamily \small
  \renewcommand{\arraystretch}{1.2}
  \caption{\textbf{Validation of Methodology.} Comparison of our regex-based license detection against a full ScanCode
  repository scan to verify coverage.}
  \label{tab:license_coverage_validation}
  \begin{tabular*}{\columnwidth}{l @{\extracolsep{\fill}} rrrr}
  \toprule
  \textbf{Metric} & \textbf{Applications} & \textbf{Models} & \textbf{Datasets} & \textbf{Combined} \\
  \midrule
  Total Repos & 380 & 364 & 345 & 1089 \\
  Perfect Match & 186 (48.9\%) & 265 (72.8\%) & 296 (85.8\%) & 747 (68.6\%) \\
  Partial Match & 137 (36.1\%) & 91 (25.0\%) & 44 (12.8\%) & 272 (25.0\%) \\
 \textbf{Any Match} & \textbf{323 (85.0\%)} & \textbf{356 (97.8\%)} & \textbf{340 (98.6\%)} & \textbf{1019 (93.6\%)} \\
  Complete Miss & 57 (15.0\%) & 8 (2.2\%) & 5 (1.4\%) & 70 (6.4\%) \\
  \midrule
  \textbf{Total} & \textbf{380} & \textbf{364} & \textbf{345} & \textbf{1089} \\
  \bottomrule
  \end{tabular*}
  \end{table}

Our license detection pipeline identifies likely license-relevant files via filename patterns (e.g., \texttt{LICENSE}, \texttt{NOTICE}, \texttt{COPYRIGHT}, and related variants), then applies ScanCode Toolkit to those retrieved files to extract license declarations and copyright notices. However, license text and notices are sometimes placed in non-standard locations (e.g., embedded in README or source files). To estimate what our retrieval stage may miss at scale, we validate its \emph{retrieval coverage} against full-repository ScanCode scans.

For a representative sample of 1089 repositories (345 datasets, 364 models, and 380 applications), we compared licenses detected by our approach against a complete ScanCode scan of all repository files. Table~\ref{tab:license_coverage_validation} reports the results. Perfect Match indicates our patterns found all licenses ScanCode detected; Partial Match indicates we found some but missed licenses in non-standard locations; Complete Miss indicates ScanCode found licenses our patterns missed entirely. Full repository scans were computationally prohibitive at our scale (38,518 AI artifacts), making pattern-based retrieval necessary. We do not run full-repository validation scans for applications.

\section{File Type Distribution in Downstream Applications}\label{app:file_types}

To inform our data collection strategy for downstream applications, we conducted an analysis of the file types present in our initial search results from the GitHub Codesearch API. Our goal was to identify the most common file extensions associated with the integration of AI models. The results of this analysis are presented in Table \ref{tab:file_distribution}.

\begin{table}[t]
\centering
\caption{Distribution of Top 15 File Types in Initial Search Results}
\label{tab:file_distribution}
\begin{tabular}{lrrr}
\toprule
\textbf{File Extension} & \textbf{Count} & \textbf{Percentage (\%)} & \textbf{\% of .py Files} \\
\midrule
.py & 1,449,523 & 37.95 & 100.00 \\
.md & 713,597 & 18.68 & 49.23 \\
.ipynb & 307,931 & 8.06 & 21.24 \\
.json & 283,296 & 7.42 & 19.54 \\
.yaml & 203,013 & 5.31 & 14.01 \\
.csv & 152,129 & 3.98 & 10.50 \\
.sh & 118,797 & 3.11 & 8.20 \\
.txt & 117,863 & 3.09 & 8.13 \\
.ts & 105,583 & 2.76 & 7.28 \\
.html & 57,628 & 1.51 & 3.98 \\
No Extension & 55,850 & 1.46 & 3.85 \\
.js & 25,092 & 0.66 & 1.73 \\
.rst & 24,770 & 0.65 & 1.71 \\
.mdx & 24,155 & 0.63 & 1.67 \\
.yml & 23,324 & 0.61 & 1.61 \\
\bottomrule
\end{tabular}
\end{table}

The data clearly shows that Python files (.py) are the predominant file type, accounting for 37.95\% of all files in the search results. The next most common executable file types, such as TypeScript (.ts) and JavaScript (.js), appear at a significantly lower frequency (2.76\% and 0.66\%, respectively). Based on this evidence, we adopted a data-driven approach by focusing our downstream analysis exclusively on Python files to capture the vast majority of AI model integrations.

While our study excludes Jupyter Notebooks (.ipynb) to focus on executable source code parsable by our AST-based approach, we acknowledge they represent a significant portion of AI-related files (8.06\%). As shown in Table \ref{tab:ipynb_stats}, including .ipynb files would increase the number of unique applications in our dataset by 27.82\%. Extending our analysis to robustly parse and analyze these notebook files is a promising direction for future work.

\begin{table}[t]
\centering
\caption{Impact of Including .ipynb Files}
\label{tab:ipynb_stats}
\begin{tabular}{lr}
\toprule
\textbf{Metric} & \textbf{Value} \\
\midrule
Unique applications with .py files & 194,544 \\
Unique applications with .py or .ipynb & 248,661 \\
Percentage increase (\%) & 27.82 \\
Additional applications from .ipynb & 54,117 \\
\bottomrule
\end{tabular}
\end{table}

\section{License Documentation Locations}
\label{app:license_locations}

Table~\ref{tab:license_locations} reports where full license text and copyright notices are located across permissively-labeled artifacts in our corpus. We find Hugging Face datasets and models store license references almost exclusively in README files (97.0\% and 95.8\%), while GitHub applications use dedicated LICENSE files (94.0\%). Copyright notices follow a similar platform divide, though they are rarer overall: only 1.9\% of datasets and 3.1\% of models include a copyright notice in a LICENSE file. These patterns support the compliance payload gap discussed in Section~\ref{subsec:license_erosion_discussion}.

\begin{table}[h]
\centering
\caption{Location of license references and copyright notices across the AI supply chain for permissively-labeled artifacts.}
\label{tab:license_locations}
\small
\begin{tabular}{@{}lrrr@{}}
\toprule
\textbf{Artifact Type} & \textbf{Total} & \textbf{LICENSE Files} & \textbf{README Files} \\
\midrule
\multicolumn{4}{@{}l}{\textit{(a) License Reference Locations}} \\
\addlinespace[2pt]
Datasets       & 1,015  &    30 (3.0\%)   &   984 (97.0\%) \\
Models         & 4,077  &   171 (4.2\%)   & 3,906 (95.8\%) \\
Applications   & 16,408 & 15,416 (94.0\%) &   992 (6.0\%)  \\
\addlinespace[2pt]
\textit{Total} & \textit{21,500} & \textit{15,617 (72.6\%)} & \textit{5,882 (27.4\%)} \\
\midrule
\multicolumn{4}{@{}l}{\textit{(b) Copyright Notice Locations}} \\
\addlinespace[2pt]
Datasets       & 1,015  &    19 (1.9\%)   &    11 (1.1\%) \\
Models         & 4,077  &   128 (3.1\%)   &    54 (1.3\%) \\
Applications   & 16,408 & 12,228 (74.5\%) &   309 (1.9\%) \\
\addlinespace[2pt]
\textit{Total} & \textit{21,500} & \textit{12,375 (57.6\%)} & \textit{374 (1.7\%)} \\
\bottomrule
\end{tabular}
\end{table}

\section{Organization Compliance}
\label{app:org_compliance}

Table~\ref{tab:org_compliance} reports license compliance for the top 15 verified Hugging Face organizations (by follower count) with permissively-labeled AI artifacts in our supply chain corpus. Of 462 artifacts across these organizations, only 7 (1.5\%) satisfy both license text and copyright notice requirements.

\begin{table}[H]
\centering
\caption{License compliance among major Hugging Face organizations. We identify verified organizations with permissively-labeled AI artifacts (MIT, Apache-2.0, BSD-3-Clause) appearing in our supply chain corpus, ranked by follower count. Has Both indicates AI artifacts with both license text and copyright notice present.}
\label{tab:org_compliance}
\resizebox{\columnwidth}{!}{%
\begin{tabular}{@{}lrrrrr@{}}
\toprule
\textbf{Organization} & \textbf{Followers} & \textbf{AI Artifacts} & \textbf{\shortstack{Has\\License}} & \textbf{\shortstack{Has\\Copyright}} & \textbf{\shortstack{Has\\Both}} \\
\midrule
Hugging Face          & 80,018  & 4   & 0.0\%  & 0.0\%   & 0.0\% \\
NVIDIA                & 49,429  & 2   & 0.0\%  & 0.0\%   & 0.0\% \\
Google                & 43,883  & 116 & 0.0\%  & 11.2\%  & 0.0\% \\
Stability AI          & 34,980  & 6   & 0.0\%  & 0.0\%   & 0.0\% \\
OpenAI                & 31,000  & 5   & 0.0\%  & 0.0\%   & 0.0\% \\
Microsoft             & 18,159  & 58  & 1.7\%  & 1.7\%   & 1.7\% \\
Unsloth AI            & 13,179  & 11  & 0.0\%  & 0.0\%   & 0.0\% \\
AI at Meta            & 11,033  & 133 & 0.0\%  & 4.5\%   & 0.0\% \\
MLX Community         & 8,433   & 9   & 0.0\%  & 0.0\%   & 0.0\% \\
Apple                 & 6,224   & 1   & 0.0\%  & 0.0\%   & 0.0\% \\
Ai2                   & 5,228   & 79  & 5.1\%  & 6.3\%   & 3.8\% \\
ByteDance             & 4,149   & 1   & 0.0\%  & 0.0\%   & 0.0\% \\
IBM Granite           & 3,889   & 13  & 0.0\%  & 0.0\%   & 0.0\% \\
Sentence Transformers & 3,540   & 9   & 0.0\%  & 0.0\%   & 0.0\% \\
Intel                 & 3,409   & 15  & 20.0\% & 20.0\%  & 20.0\% \\
\midrule
\textbf{Total}        &         & \textbf{462} & \textbf{1.7\%} & \textbf{6.3\%} & \textbf{1.5\%} \\
\bottomrule
\end{tabular}
}
\end{table}

\section{Traceability and Documentation Artifacts}
\label{app:traceability_artifacts}

Table~\ref{tab:lineage_disclosure} reports lineage disclosure practices across all 1,922,385 Hugging Face models in our initial corpus. Only 7.1\% declare training datasets via metadata, which bounds the supply chains available for end-to-end auditing. Table~\ref{tab:missing_artifacts} reports the availability of LICENSE and README files across our full supply chain corpus of 38,518 artifacts. LICENSE files are nearly absent upstream (96.5\% of datasets and 93.4\% of models), while README files show the inverse pattern. 

\begin{table}[H]
\centering
\small
\caption{Lineage disclosure practices across Hugging Face models. Only 7.1\% of models declare training datasets, limiting supply chain traceability.}
\label{tab:lineage_disclosure}
\begin{tabular}{lrr}
\toprule
\textbf{Metadata Field} & \textbf{Count} & \textbf{\% of Total} \\
\midrule
Total Models & 1,922,385 & 100.0\% \\
\midrule
Has \texttt{base\_model} tag & 568,496 & 29.6\% \\
Has \texttt{datasets} tag & 136,890 & 7.1\% \\
Has at least one like & 148,729 & 7.7\% \\
\bottomrule
\end{tabular}
\end{table}

\begin{table}[H]
\centering
\small
\caption{Availability of LICENSE/README across the AI supply chain. Missing LICENSE indicates no LICENSE file detected; Missing README indicates no README file detected.}
\label{tab:missing_artifacts}
\begin{tabular}{lrrrr}
\toprule
\textbf{Artifact} & \textbf{Total} & \textbf{\shortstack{Missing\\README}} & \textbf{\shortstack{Missing\\LICENSE}} & \textbf{\shortstack{Missing\\Either}} \\
\midrule
Datasets & 3,338 & 1,158 (34.7\%) & 3,222 (96.5\%) & 3,237 (97.0\%) \\
Models & 6,664 & 1,224 (18.4\%) & 6,221 (93.4\%) & 6,278 (94.2\%) \\
Apps & 28,516 & 18,973 (66.5\%) & 11,357 (39.8\%) & 20,398 (71.5\%) \\
\midrule
\textbf{Total} & \textbf{38,518} & \textbf{21,355 (55.4\%)} & \textbf{20,800 (54.0\%)} & \textbf{29,913 (77.7\%)} \\
\bottomrule
\end{tabular}
\end{table}

\section{Compliance-Relevant File Selection Patterns}
\label{app:file_patterns}

To retrieve compliance-relevant files from AI artifact repositories at scale, we define two sets of case-insensitive regex patterns: one for license-related files and one for README files. These patterns are applied to every file path in a repository to determine which files to download for license text extraction and copyright notice detection. We impose a global file size limit of 300\,MB per file to exclude binary artifacts (e.g., model weights, dataset archives) that are unlikely to contain license text.

\subsection{License File Patterns}

We target files whose names or directory paths suggest license, copyright, or notice content. Our keyword list extends conventions from prior large-scale license studies~\citep{Zacchiroli_2022} to cover AI-specific naming patterns (e.g., \texttt{model\_license}, \texttt{datasheet}):

\begin{small}
\begin{verbatim}
LICENSE_KEYWORDS = [
  'license', 'licence', 'copying',
  'unlicense', 'patents', 'notice',
  'copyright', 'disclaimer', 'authors',
  'legal', 'terms', 'attribution',
  'citation', 'model_license',
  'data_license', 'dataset_license',
  'modelcard', 'model_card', 'datasheet'
]

LICENSE_DIRECTORIES = [
  'legal', 'license', 'licenses',
  'licensing', 'copyright', 'terms',
  'attribution'
]
\end{verbatim}
\end{small}

The license file regex matches any file whose basename contains a keyword, optionally preceded by a prefix or followed by an extension:

\begin{small}
\begin{verbatim}
LICENSE_REGEX =
  r"^(.*\/)?([a-z0-9._-]+\.)?("
  + "|".join(LICENSE_KEYWORDS)
  + r")(\.[a-z0-9\._-]+)?$"
\end{verbatim}
\end{small}

\noindent This pattern is applied with the case-insensitive flag and matches paths such as \texttt{LICENSE}, \texttt{LICENSE.md}, and \texttt{pkg.license.txt}. We classify matched files into three categories based on their location within the repository:

\begin{itemize}
    \item \textbf{Root license}: a license-relevant file in the repository root (e.g., \texttt{LICENSE}, \texttt{NOTICE.md}).
    \item \textbf{Directory license}: any file located inside a directory matching \texttt{LICENSE\_DIRECTORIES} (e.g., all files under \texttt{legal/}). We retain \emph{all} files within matched directories, as they often contain supplementary notices, third-party attributions, or license variants that collectively form the compliance payload.
    \item \textbf{Scattered license}: a license-relevant file located elsewhere in the repository tree (e.g., \texttt{src/third\_party/LICENSE}).
\end{itemize}

\subsection{README File Patterns}

We also retrieve README files because on Hugging Face, the conventional location for YAML metadata, including license declarations~\cite{huggingface_hub_model_cards_guide}. We match README files using:

\begin{small}
\begin{verbatim}
README_PATTERNS = [
  'readme', 'read_me', 'read-me'
]

README_REGEX =
  r"^(.*\/)?("
  + "|".join(README_PATTERNS)
  + r")(\..*)?$"
\end{verbatim}
\end{small}

\noindent This pattern matches files such as \texttt{README.md}, \texttt{READ\_ME.rst}, and \texttt{docs/readme.txt}.

Both regex patterns are applied with the case-insensitive flag (\texttt{re.IGNORECASE}) to account for inconsistent capitalization across platforms. We validate the retrieval coverage of these patterns against full-repository ScanCode scans in Appendix~\ref{app:license_retrieval_validation}.

\section{Replication Package}
\label{app:replication}

Our audit dataset and reproducible pipeline are publicly available at: \url{https://github.com/nn8441731-lgtm/nn8441731}.
\end{document}